\documentclass[10pt,twocolumn,letterpaper]{article}

\usepackage{cvpr}
\usepackage{times}
\usepackage{epsfig}
\usepackage{graphicx}
\usepackage{amsmath}
\usepackage{amssymb}
\usepackage{afterpage}

\usepackage[pagebackref=true,breaklinks=true,letterpaper=true,colorlinks,bookmarks=false]{hyperref}

\cvprfinalcopy 

\def\cvprPaperID{3335} 

\ifcvprfinal\fi
\begin{document}

\title{Image to Image Translation for Domain Adaptation}

\author{{Zak Murez$^{1,2}$} \quad {Soheil Kolouri$^ {2}$} \quad {David Kriegman$^1$} \quad  {Ravi Ramamoorthi$^1$} \quad {Kyungnam Kim$^ {2}$}\\
{$^1$ University of California, San Diego;}
{$^2$ HRL Laboratories, LLC;} \\
{\tt\small { {\{zmurez,kriegman,ravir\}}@cs.ucsd.edu}, \{skolouri,kkim\}}@hrl.com
\vspace{-2mm}
}

\maketitle

\begin{abstract}
We propose a general framework for unsupervised domain adaptation, which allows deep neural networks trained on a source domain to be tested on a different target domain without requiring any training annotations in the target domain. This is achieved by adding extra networks and losses that help regularize the features extracted by the backbone encoder network. To this end we propose the novel use of the recently proposed unpaired image-to-image translation framework to constrain the features extracted by the encoder network. Specifically, we require that the features extracted are able to reconstruct the images in both domains. In addition we require that the distribution of features extracted from images in the two domains are indistinguishable. Many recent works can be seen as specific cases of our general framework. We apply our method for domain adaptation between MNIST, USPS, and SVHN datasets, and Amazon, Webcam and DSLR Office datasets in classification tasks, and also between GTA5 and Cityscapes datasets for a segmentation task. We demonstrate state of the art performance on each of these datasets.

\end{abstract}

\section{Introduction}

\begin{figure}[!!t]
\centering
\includegraphics[width=.99\columnwidth]{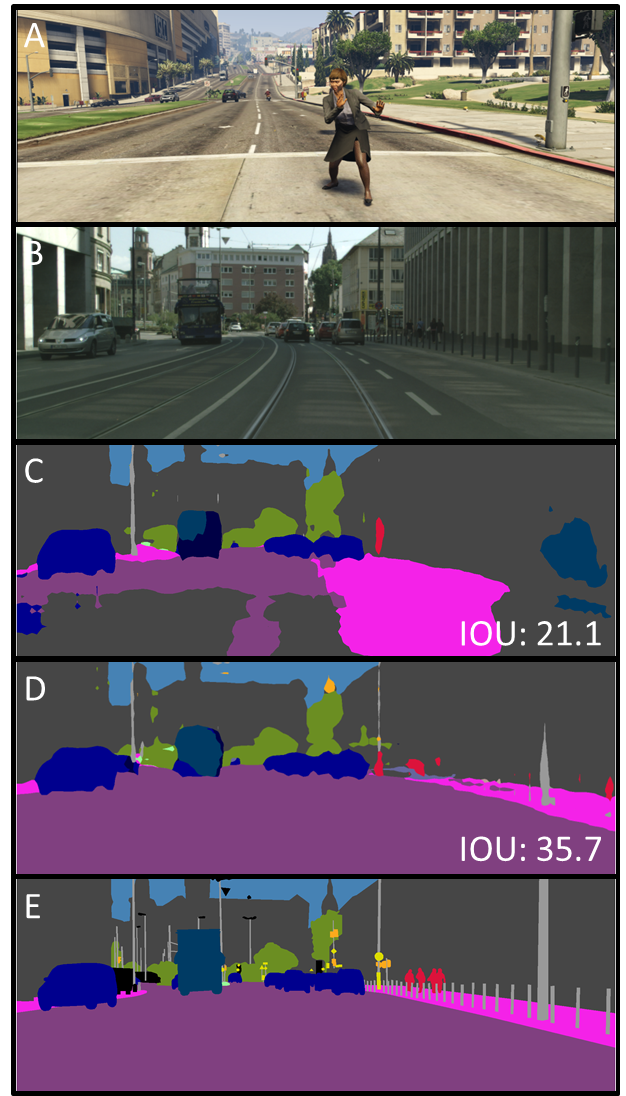}
\caption{A) Sample image from the synthetic GTA5 dataset. B) Input image from the real Cityscapes dataset. C) Segmentation result trained on GTA5 dataset without any domain adaptation. D) Ours. E) Ground truth. We can see that our adaptation fixes large areas of simple mistakes on the road and sidewalk and building on the right. We also partially detect the thin pole on the right. The mean Intersection Over Union (IOU) values are reported. 
}
\label{fig:teaser}\vspace{-0.5cm}
\end{figure}

The recent unprecedented advances in computer vision and machine learning are mainly due to: 1) deep (convolutional) neural architectures, and 2) existence of abundant labeled data. Deep convolutional neural networks (CNNs) \cite{krizhevsky2012imagenet,he2016deep,huang2016densely} trained on large numbers of labeled images (tens of thousands to millions) provide powerful image representations that can be used for a wide variety of tasks including recognition, detection, and segmentation. On the other hand, obtaining abundant annotated data remains a cumbersome and expensive process in the majority of applications. Hence, there is a need for transferring the learned knowledge from a source domain with abundant labeled data to a target domain where data is unlabeled or sparsely labeled. The major challenge for such knowledge transfer is a phenomenon known as domain shift \cite{gretton2009covariate}, which refers to the different distribution of data in the target domain compared to the source domain. 

To further motivate the problem, consider the emerging application of autonomous driving where a semantic segmentation network is required to be trained to detect roads, cars, pedestrians, etc. Training such segmentation networks requires semantic, instance-wise, dense pixel annotations for each scene, which is excruciatingly expensive and time consuming to acquire. To avoid human annotations, a large body of work focuses on designing photo-realistic simulated scenarios in which the ground truth annotations are readily available. Synthia~\cite{RosCVPR16}, Virtual KITTI~\cite{gaidon2016virtual}, and GTA5~\cite{richter2016playing} datasets are examples of such simulations, which  include a large number of synthetically generated driving scenes together with ground truth  pixel-level semantic annotations. Training a CNN based on such synthetic data and applying it to real-world images (i.e. from a dashboard mounted camera), such as the Cityscapes dataset~\cite{Cordts2016Cityscapes}, will give very poor performance due to the large differences in image characteristics which gives rise to the domain shift problem. Figure \ref{fig:teaser} demonstrates this scenario where a network is trained on the GTA5 dataset \cite{richter2016playing}, which is a synthetic dataset, for semantic segmentation and is tested on the Cityscapes dataset \cite{Cordts2016Cityscapes}. It can be seen that with no adaptation the network struggles with segmentation (Figure \ref{fig:teaser}, C), while our proposed  framework ameliorates the domain shift problem and provides a more accurate semantic segmentation.  


Domain adaptation techniques aim to address the domain shift problem, by finding a mapping from the source data distribution to the target distribution. Alternatively, both domains could be mapped into a shared domain where the distributions are aligned. Generally, such mappings are not unique and there exist many mappings that align the source and target distributions. Therefore various constraints are needed to narrow down the space of feasible mappings. Recent domain adaptation techniques parameterize and learn these mappings via deep neural networks \cite{tzeng2015simultaneous,long2015learning,tzeng2017adversarial,luo_nips17_label}.
In this paper, we propose a unifying, generic, and systematic framework for unsupervised domain adaptation, which 
is broadly applicable to many image understanding and sensing  tasks where training labels are not available in the target domain. We further demonstrate that many existing methods for domain adaptation arise as special cases of our framework.



While there are significant differences between the recently developed domain adaptation methods, a common and unifying theme among these methods can be observed. We identify three main attributes needed to achieve successful unsupervised domain adaptation: 1) domain agnostic feature extraction, 2) domain specific reconstruction, and 3) cycle consistency.  
The first requires that the distributions of features extracted from both domains are indistinguishable (as judged by an adversarial discriminator network). This idea was utilized in many prior methods \cite{hoffman2016fcns,ganin2015unsupervised,ganin2016domain}, but alone does not give a strong enough constraint for domain adaptation knowledge transfer, as there exist many mappings that could match the source and target distributions in the shared space. The second is requiring that the features are able to be decoded back to the source and target domains.  This idea was used in Ghifary et al.~\cite{ghifary2016deep} for unsupervised domain adaptation. Finally, the cycle consistency is needed for unpaired source and target domains to ensure that the mappings are learned correctly and they are well-behaved, in the sense that they do not collapse the distributions into single modes \cite{zhu2017unpaired}. Figure \ref{fig:hlevel} provides a high-level overview of our framework. 

The interplay between the `domain agnostic feature extraction', `domain specific reconstruction with cycle consistency', and `label prediction from agnostic features' enables our framework to simultaneously learn from the source domain and adapt to the target domain. By combining all these different components into a single unified framework we build a systematic framework for domain knowledge transfer that provides an elegant theoretical explanation as well as improved experimental results. We demonstrate the superior performance of our proposed framework for segmentation adaptation from synthetic images to real world images (See Figure \ref{fig:teaser} as an example), as well as for classifier adaptation on three digit datasets. Furthermore, we show that many of the State Of the Art (SOA) methods can be viewed as special cases of our proposed framework.

\section{Related Work}
\begin{figure}
\centering
\includegraphics[width=\columnwidth]{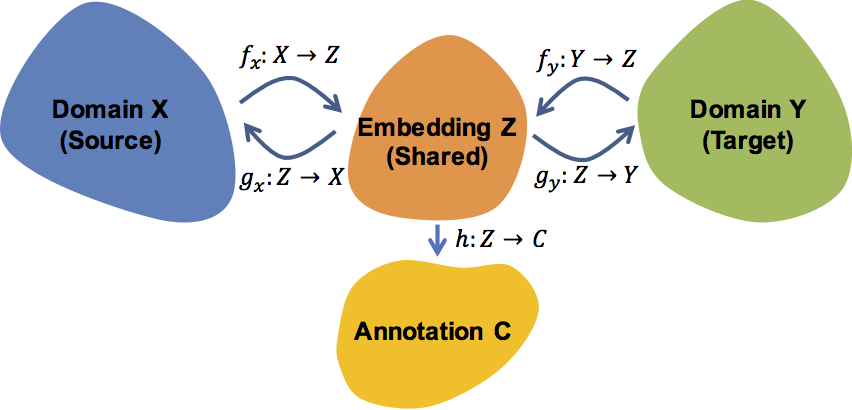}
\caption{The source, target, annotation, and shared embedding spaces with the corresponding mappings between them.}
\label{fig:hlevel}
\end{figure}

There has been a plethora of recent work in the field of visual {\it domain adaptation} addressing the domain shift problem \cite{gretton2009covariate}, otherwise known as the {\it dataset  bias problem}. The majority of recent work use deep convolutional architectures to map the source and target domains into a shared space where the domains are aligned \cite{tzeng2014deep,tzeng2015simultaneous,tzeng2017adversarial,hoffman2016fcns}.
These methods widely differ on the architectures as well as the choices of loss functions used for training them. Some have used Maximum Mean Discrepancy (MMD) between the distributions of the source and target domains in the shared space \cite{long2015learning}, while others have used correlation maximization to align the second-order statistics of the domains. Another popular and effective choice is maximizing the confusion rate of an adversarial network, that is required to distinguish the source and target domains in the shared space \cite{tzeng2015simultaneous,hoffman2016fcns,ganin2015unsupervised,ganin2016domain,ghifary2016deep}. Other approaches include the work by Sener et al. \cite{sener2016learning}, where the domain transfer is formulated in a transductive setting, and the Residual Transfer Learning (RTL) approach \cite{long2016unsupervised} where the authors assume that the source and target classifiers only differ by a residual function and learn these residual functions.

Our work is primarily motivated by the work of Hoffman et al. \cite{hoffman2016fcns}, Isola et al. \cite{isola2016image}, Zhu et al. \cite{zhu2017unpaired}, and Ghifary et al. \cite{ghifary2016deep}. Hoffman et al. \cite{hoffman2016fcns}  
utilized fully convolutional networks with domain adversarial training to obtain domain agnostic features (i.e. shared space) for the source and target domains, while constraining the shared space to be discriminative for the source domain. Hence, by learning the mappings from source and target domains to the shared space (i.e. $f_x$ and $f_y$ in Figure \ref{fig:hlevel}), and learning the mapping from the shared space to annotations (i.e. $h$ in Figure \ref{fig:hlevel}), their approach effectively enables the learned classifier to be applicable to both domains. The Deep Reconstruction Classification
Network (DRCN) of Ghifary et al. \cite{ghifary2016deep}, utilizes a similar approach but with a constraint that the embedding must be decodable, and learns a mapping from the embedding space to the target domain (i.e. $g_y$ in Figure \ref{fig:hlevel}). The image-to-image translation work by Isola et al. \cite{isola2016image} maps the source domain to the target domain by an adversarial learning of $f_x$ and $g_y$ and composing them $g_y\circ f_x:X\rightarrow Y$. In their framework the target and source images were assumed to be paired, in the sense that for each source image there exists a known corresponding target image. This assumption was lifted in the follow-up work of Zhu et al. \cite{zhu2017unpaired}, where cycle consistency was used to learn the mappings based on unpaired source and target images. While the approaches of Isola et al. \cite{isola2016image} and Zhu et al. \cite{zhu2017unpaired} do not address the domain adaptation problem, yet they provide a baseline for learning high quality mappings from a visual domain into another. 

The patterns that collectively emerge from the mentioned papers \cite{tzeng2014deep,hoffman2016fcns,isola2016image,ghifary2016deep,zhu2017unpaired}, are: a) the shared space must be a discriminative embedding for the source domain, b) the embedding must be domain agnostic, hence maximizing the similarity between the distributions of embedded source and target images, c) the information preserved in the embedding must be sufficient for reconstructing domain specific images, d) adversarial learning as opposed to the classic losses can significantly enhance the quality of learned mappings, e) cycle-consistency is required to reduce the space of possible mappings and ensure their quality, when learning the mappings from unpaired images in the source and target domains. Our proposed method for unsupervised domain adaptation unifies the above-mentioned pieces into a generic framework that simultaneously solves the domain adaptation and image-to-image translation problems.

There have been other recent efforts toward a unifying and general framework for deep domain adaptation. The Adversarial Discriminative Domain Adaptation (ADDA) work by Tzeng et al. \cite{tzeng2017adversarial} 
is an instance of such frameworks. Tzeng et al. \cite{tzeng2017adversarial} identify three design choices for a deep domain adaptation system, namely a) whether to use a generative or discriminative base, whether to share mapping parameters between $f_x$ and $f_y$, and the choice of adversarial training. They observed that modeling image distributions might not be strictly necessary if the embedding is domain agnostic (i.e. domain invariant). 

\section{Method}
\begin{figure*}[t]
\centering
\includegraphics[width=0.99\linewidth]{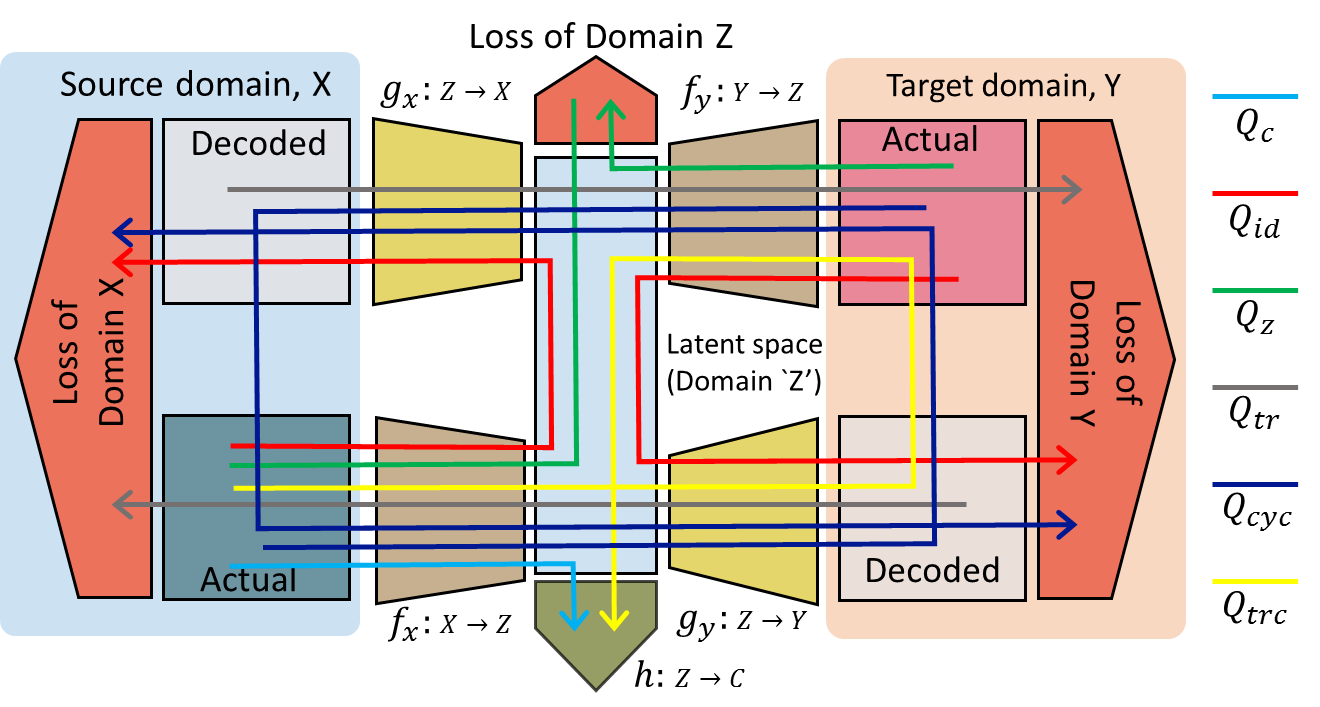}
\caption{The detailed system architecture of our I2I (image to image) Adapt framework. The pathways to the loss modules denote the inputs to these modules, which are used for training. Best viewed in color.}
\label{fig:model}
\end{figure*}

Consider training images $x_i\in X$ and their corresponding annotations/labels $c_i\in C$ from the source domain (i.e. domain $X$). Note that $c_i$ may be image level such as in classification or pixel level in the case of semantic segmentation. Also consider training images $y_j\in Y$ in the target domain (i.e. domain $Y$), where we do not have corresponding annotations for these images. 
Our goal is then to learn a classifier that maps the target images, $y_j$s, to labels $C$. We note that the framework is readily extensible to a semi-supervised learning or few-shot learning scenario where we have annotations for a few images in the target domain. Given that the target domain lacks labels, the general approach is to learn a classifier on the source domain and adapt it in a way that its domain distribution matches that of the target domain. 

The overarching idea here is to find a joint latent space, $Z$, for the source and target domains, $X$ and $Y$, where the representations are domain agnostic. To clarify this point, consider the scenario in which $X$ is the domain of driving scenes/images on a sunny day and $Y$ is the domain of driving scenes on a rainy day. While `sunny' and `rainy' are characteristics of the source and target domains, they are truly nuisance variations with respect to the annotation/classification task (e.g. semantic segmentation of the road), as they should not affect the annotations. Treating such characteristics as structured noise, we would like to find a latent space, $Z$, that is invariant to such variations. In other words, domain $Z$ should not contain domain specific characteristics, hence it should be domain agnostic. In what follows we describe the process that leads to finding such a domain agnostic latent space. 
 
Let the mappings from source and target domains to the latent space be defined as $f_x:X\rightarrow Z$ and $f_y:Y\rightarrow Z$, respectively (See Figure \ref{fig:hlevel}). 
In our framework these mappings are parameterized by deep convolutional neural networks (CNNs). Note that the members of the latent space $z\in Z$ are high dimensional vectors in the case of image level tasks, or feature maps in the case of pixel level tasks. Also, let $h:Z\rightarrow C$ be the classifier that maps the latent space to labels/annotations (i.e. the classifier module in Figure \ref{fig:model}). Given that the annotations for the source class $X$ are known, one can define a supervised loss function to enforce $h(f_x(x_i))=c_i$:
\begin{equation} 
Q_c=\sum_i l_c\left(h(f_x(x_i)), c_i\right)
\end{equation}
where $l_c$ is an appropriate loss (e.g. cross entropy for classification and segmentation). Minimizing the above loss function leads to  the standard approach of supervised learning, which does not concern domain adaptation. While this approach would lead to a method that performs well on the images in the source domain, $x_i\in X$, it will more often than not perform poorly on images from the target domain $y_j\in Y$. The reason is that, domain $Z$ is biased to the distribution of the structured noise (`sunny') in domain $X$ and the structured noise in domain $Y$ (`rainy') confuses the classifier $h(\cdot)$. To avoid such confusion we require the latent space, $Z$, to be domain agnostic, so it is not sensitive to the domain specific structured noise. To achieve such a latent space we systematically introduce a variety of auxiliary networks and losses to help regularize the latent space and consequently achieve a robust $h(\cdot)$. The auxiliary networks and loss pathways are depicted in Figure \ref{fig:model}. In what follows we describe the individual components of the regularization losses.

\begin{enumerate}
\item First of all $Z$ is required to preserve the core information of the target and source images and only discard the structured noise. To impose this constraint on the latent space, we first define decoders $g_x:Z\rightarrow X$ and $g_y:Z\rightarrow Y$ that take the features in the latent space to the source and target domains, respectively. We assume that if $Z$ retains the crucial/core information of the domains and only discards the structured noise, then the decoders should be able to add the structured noise back and reconstruct each image from their representation in the latent feature space,  $Z$. In other words, we require $g_x(f_x(\cdot))$ and $g_y(f_y(\cdot))$ to be close to identity functions/maps. This constraint leads to the following loss function:
\begin{align}
Q_{id}=
&\sum_i l_{id}\left(g_x(f_x(x_i)),x_i\right)
+\nonumber\\
&\sum_j l_{id}\left(g_y(f_y(y_j)),y_j\right)
\end{align}
where $l_{id}(\cdot,\cdot)$ is a pixel-wise image loss such as the $L_1$ norm. 

\item We would like the latent space $Z$ to be domain agnostic. This means that the feature representations of the source and target domain should not contain domain specific information. To achieve this, we use an adversarial setting in which a discriminator $d_z: Z \rightarrow \{c_x,c_y\}$ tries to classify if a feature in the latent space $z\in Z$ was generated from domain $X$ or $Y$, where $c_x$ and $c_y$ are binary domain labels (i.e. from domain X or domain Y). The loss function then can be defined as the certainty of the discriminator (i.e. domain agnosticism is equivalent to fooling the discriminator), and therefore we can formulate this as:
\begin{align}
Q_{z}=
&\sum_i l_a\left(d_z(f_x(x_i)),c_x\right) +\nonumber\\
&\sum_j l_{a}\left(d_z(f_y(y_j)),c_y\right)
\end{align}
where $l_a(\cdot,\cdot)$ is an appropriate loss (the cross entropy loss in traditional GANs~\cite{goodfellow2014generative} and mean square error in least squares GAN~\cite{mao2016multi}). The discriminator is trained to maximize this loss while the discriminator is trained to minimize it. 

\item To further ensure that the mappings $f_x$, $f_y$, $g_x$, and $g_y$ are consistent we define translation adversarial losses. An image from target (source) domain is first encoded to the latent space and then decoded to the source (target) domain to generate a `fake' (translated) image. Next, we define discriminators $d_x:X\rightarrow \{c_x,c_y\}$ and $d_y:Y\rightarrow\{c_x,c_y\}$, to identify if an image is `fake' (generated from the other domain) or `real' (belonged to the actual domain). To formulate this translation loss function we can write:
\begin{align}
Q_{tr}=
&\sum_i l_{a}\left(d_y(g_y(f_x(x_i))),c_x\right)
+\nonumber\\
&\sum_j l_a\left(d_x(g_x(f_y(y_j)),c_y\right)
\end{align}

\item Given that there are no correspondences between the images in the source and target domains, we need to ensure that the semantically similar images in both domains are projected into close vicinity of one another in the latent space. To ensure this,  we define the cycle consistency losses where the `fake' images generated in the translation loss, $g_x(f_y(y_j))$ or $g_y(f_x(x_i))$, are encoded back to the latent space and then decoded back to their original space. The entire cycle should be equivalent to an identity mapping. We can formulate this loss as follows:
\begin{align}
Q_{cyc}=
&\sum_i l_{id}\left(g_x(f_y(g_y(f_x(x_i)))),x_i\right)
+\nonumber\\
&\sum_j l_{id}\left(g_y(f_x(g_x(f_y(y_j)))),y_j\right)
\end{align}


\item To further constrain the translations to maintain the same semantics, and allow the target encoder to be trained with supervision on target domain `like' images we also define a classification loss between the source to target translations and the original source labels:
\begin{equation} 
Q_{trc}=\sum_i l_c\left(h(f_y(g_y(f_x(x_i)))), c_i\right)
\end{equation}

\end{enumerate}
Finally, by combining these individual losses we define the general loss to be,
\begin{equation}
Q =  \lambda_c Q_c 
   + \lambda_{z}Q_{z}
   + \lambda_{tr}Q_{tr}
   + \lambda_{id}Q_{id}
   + \lambda_{cyc}Q_{cyc}
   + \lambda_{trc}Q_{trc}
\end{equation}
The above general loss function is then optimized via Stochastic Gradient Descent (SGD) method with adaptive learning rate, in an end-to-end manner. Figure \ref{fig:model} shows the pathways for each loss function defined above. The discriminative networks, $d_x$, $d_y$, and $d_z$ are trained in an alternating optimization alongside with the encoders and decoders. 

To further constrain the features that are learned we share the weights of the encoders. We also share the weights of the first few layers of the decoders. To stabilize the image domain discriminators we train them using the Improved Wasserstein method~\cite{gulrajani2017improved}. We found that the Wasserstein GAN is not well suited for discriminating in the Z domain since both the `real' and `fake' distributions are changing. As such we resort to using the least squares GAN~\cite{mao2016multi} for the Z domain.

 Here we will show how various previous methods for domain adaptation are special cases of our method. By setting $\lambda_{id}=\lambda_{cyc}=\lambda_{tr}=0$ we recover~\cite{hoffman2016fcns}. By first training only on the source domain and then freezing the source encoder, untying the target encoder and setting $\lambda_{id}=\lambda_{cyc}=\lambda_{tr}=0$ we recover~\cite{tzeng2017adversarial}. By setting $\lambda_{id_A}=\lambda_{cyc}=\lambda_{tr}=\lambda_{z}=0$  we recover~\cite{ghifary2016deep}, where $\lambda_{id_A}$ indicates the mixing coefficient only for the first term of $Q_{id}$. 
 Finally, by setting $\lambda_{id}=\lambda_c=\lambda_{z}=0$ we recover~\cite{zhu2017unpaired}. Table \ref{tab:soa} summarizes these results. 

\begin{table}[t]
\small
\centering
\begin{tabular}{ c c c c c c c c} 
\hline
Method & $\lambda_c$ & $\lambda_z$ & $\lambda_{tr}$ & $\lambda_{id_A}$ & $\lambda_{id_B}$  &  $\lambda_{cyc}$ & $\lambda_{trc}$ \\
\hline
\cite{hoffman2016fcns}  & \checkmark & \checkmark & \\
\cite{tzeng2017adversarial} & \checkmark & \checkmark & \\
\cite{ghifary2016deep} & \checkmark & & & & \checkmark & &\\
\cite{zhu2017unpaired} & & &\checkmark & & &\checkmark &
\\
\hline
Ours & \checkmark &\checkmark &\checkmark &\checkmark &\checkmark &\checkmark &\checkmark 
\\
\hline
\end{tabular}
\caption{Showing the relationship between the existing methods and our proposed method.}
\label{tab:soa}
\end{table}

\section{Experiments} 
\subsection{MNIST, USPS, and SVHN digits datasets}
\begin{figure*}
\centering
\includegraphics[width=.955\linewidth]{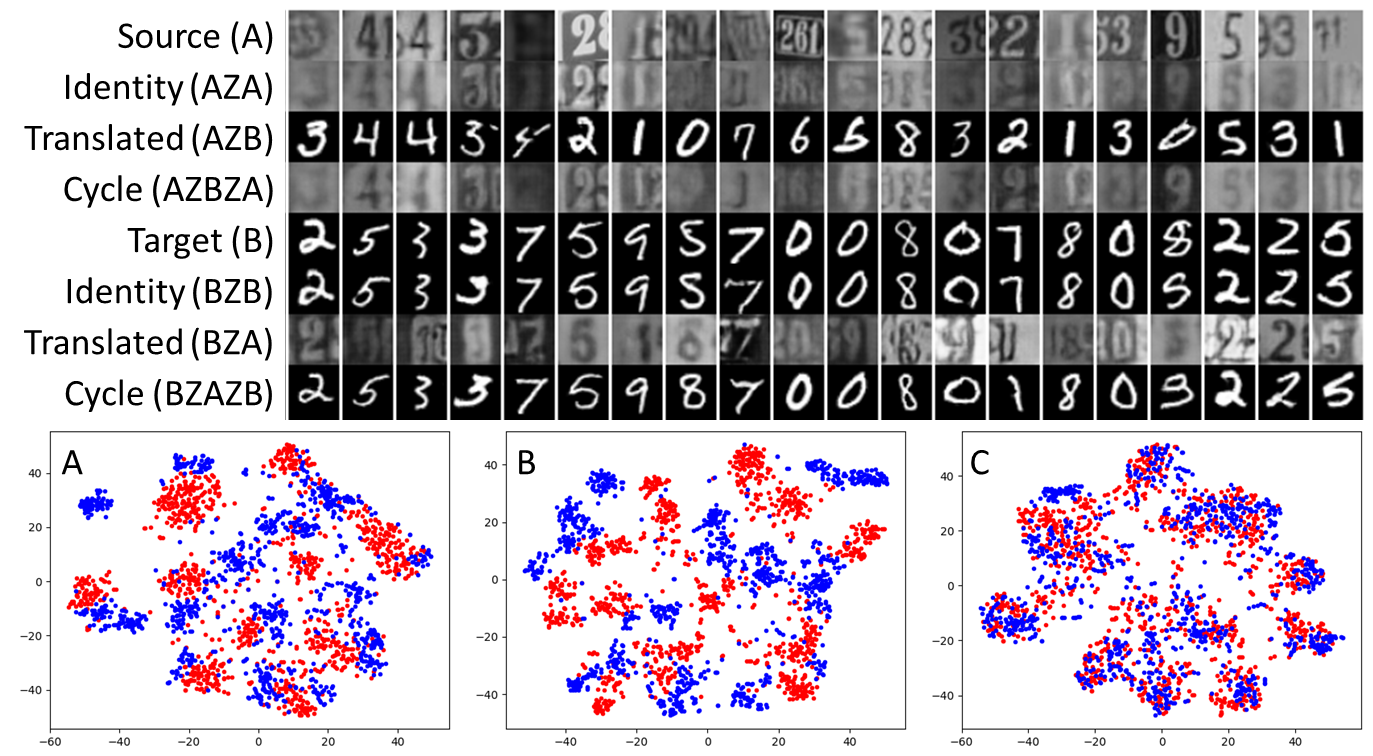}
\caption{Top) Image to image translation examples for MNIST to SVHN. Bottom) TSNE embedding visualization of the latent space. Red are source images, Blue are target images. A) No adaptation. B) Image to image adaptation without latent space discriminator. C) Full adaptation.}
\label{fig:digits}
\end{figure*}

First, we demonstrate our method on domain adaptation between three digit classification datasets, namely MNIST \cite{lecun1998gradient}, USPS \cite{hull1994database}, and the Street View House Numbers (SVHN) \cite{netzer2011reading} datasets. The MNIST dataset consists of 60,000 training and 10,000 test binary images of handwritten digits of size $28\times 28$. The USPS dataset contains 7291 training and 2007 test grayscale images of handwritten images of size $16\times 16$. The SVHN dataset, which is a significantly more challenging dataset, contains 73,257 training and 26,032 digits test RGB images of size $32\times 32$. We performed the same experiments as in \cite{ganin2016domain,tzeng2015simultaneous,liu2016coupled,tzeng2017adversarial}
where we treated one of the digit datasets as a labeled source domain and another dataset as unlabeled target domain. We trained our framework for adaptation from MNIST$\rightarrow$ USPS, USPS $\rightarrow$ MNIST, and SVHN $\rightarrow$ MNIST. Figure \ref{fig:digits} shows examples of MNIST to SVHN input and translated images. 

For a fair comparison with previous methods, our feature extractor network (encoder, $f_x$ and $f_y$) is a modified version of LeNet~\cite{lecun1998gradient}. Our decoders (i.e. $g_x$ and $g_y$) consist of three transposed convolutional layers with batch normalization and leaky ReLU nonlinearities. Our image discriminators consist of three convolutional layers and our feature discriminator consists of three fully connected layers. We also experimented with a deeper DenseNet architecture~\cite{huang2016densely} for the encoder which improved performance for all methods (in fact DenseNet without any domain adaptation beat almost all prior methods that include domain adaptation). We compare our method to five prior works (see Table. \ref{table:digits}). Our method consistently out performs prior work, and when combined with the DenseNet architecture, significantly outperforms the prior SOA.


Figure \ref{fig:digits} A,B,C show TSNE embeddings of the features extracted from the source and target domain when trained without adaptation, with image to image loss only, and our full model. It can be seen that without adaptation, the source and target images get clustered in the feature space but the distributions do not overlap which is why classification fails on the target domain. Just image to image translation is not enough to force the distributions to overlap as the networks learn to map source and target distributions to different areas of the feature space. Our full model includes a feature distribution adversarial loss, forcing the  source and target distributions to overlap, while image translation makes the features richer yielding the best adaptation results.

\begin{table*}
\small
\centering
\begin{tabular}{ c c c c } 
\hline
Method & MNIST $\rightarrow$ USPS & USPS $\rightarrow$ MNIST & SVHN $\rightarrow$ MNIST \\ 
\hline
Source only & 75.2 & 57.1 & 60.1\\ 
Gradient reversal~\cite{ganin2016domain} & 77.1 & 73.0 & 73.9\\ 
Domain confusion~\cite{tzeng2015simultaneous} & 79.1 & 66.5 & 68.1\\ 
CoGAN~\cite{liu2016coupled} & 91.2 & 89.1 & -\\ 
ADDA~\cite{tzeng2017adversarial} & 89.4 &\textcolor{blue}{ 90.1} & 76.0 \\ 
I2I Adapt (Ours) & \textcolor{blue}{92.1} & 87.2 & \textcolor{blue}{80.3}\\ 
\hline
Source only - DenseNet & 95.0 & 88.1 & 80.1\\ 
I2I Adapt - DenseNet (Ours) & \bf{95.1} & \bf{92.2} & \bf{92.1}\\ 
\hline
\end{tabular}
\caption{Performance of various methods on digits datasets domain adaptation. MNIST $\rightarrow$ USPS indicates MNIST is the source domain (labels available) and USPS is the target domain (no labels available). Source only is the baseline no domain adaptation. Above the line uses the standard LeNet architecture for the encoder. 
Below the line we have replaced the encoder with the recent DenseNet architecture. DenseNet without domain adaptation beats almost all previous methods and when combined with our method beats all previous methods by a significant margin. Blue is best with LeNet, bold is best overall.}
\label{table:digits}
\end{table*}

\subsection{Office dataset}

\begin{figure}[h]
\centering
\includegraphics[width=.9\linewidth]{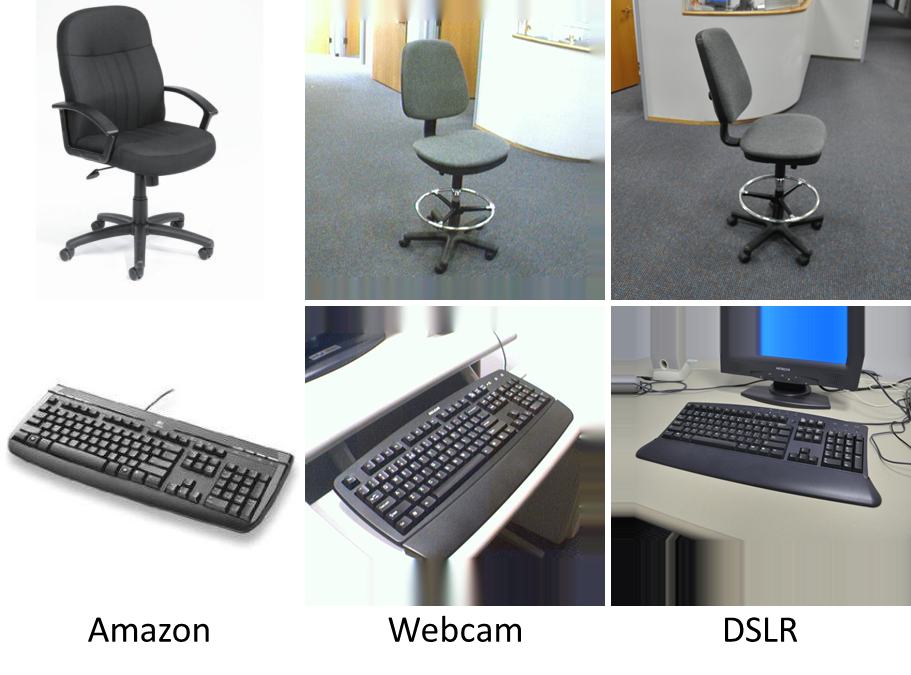}
\caption{Sample images from the Office dataset}
\label{fig:sample}
\end{figure}

The Office dataset~\cite{saenko2010adapting} consists of images from 31 classes of objects in three domains: Amazon (A), Webcam (W) and DSLR (D) with 2817, 795 and 498 images respectively (see Figure~\ref{fig:sample} for examples). Our method performs the best in four out of six of the tasks (see Table~\ref{table:office}). The two tasks that ours is not best at consist of bridging a large domain shift with very little training data in the source domain (795 and 498 respectively).

\begin{table*}
\small
\centering
\begin{tabular}{ c c c c c c c } 
\hline
Method & A $\rightarrow$ W & W $\rightarrow$ A & A $\rightarrow$ D & D $\rightarrow$ A & W $\rightarrow$ D & D $\rightarrow$ W \\ 
\hline
Domain confusion~\cite{tzeng2014deep} & 61.8 & 52.2 & 64.4 & 21.1 & 98.5 & 95.0 \\ 
Transferable Features~\cite{long2015learning} & 68.5 & 53.1 & 67.0 & 54.0 & 99.0 & 96.0  \\ 
Gradient reversal~\cite{ganin2016domain} & 72.6 & 52.7 & 67.1 & 54.5 & 99.2 & 96.4 \\ 
Reconstruction~\cite{ghifary2016deep} & 68.7 & \bf{54.9} & 66.8 & \bf{56.0} & 99.0 & 96.4 \\ 
I2I Adapt (Ours) & \bf{75.3} & 52.1 & \bf{71.1} & 50.1 & \bf{99.6} & \bf{96.5} \\ 
\hline
\end{tabular}
\caption{Performance of various methods on the Office dataset consisting of three domains: Amazon (A), Webcam (W) and DSLR (D).  A~$\rightarrow$~W indicates Amazon is the source domain (labels available) and Webcam is the target domain (no labels available). Bold is best. Our method performs best on 4 out of 6 of the tasks. }
\label{table:office}
\end{table*}

\subsection{GTA5 to Cityscapes}

We also demonstrate our method for domain adaptation between the synthetic (photorealistic) driving dataset GTA5~\cite{richter2016playing} and the real dataset Cityscapes~\cite{Cordts2016Cityscapes}. The GTA5 dataset consists of 24,966 densely labeled RGB images (video frames) of size $1914\times 1052$, containing 19 classes that are compatible with the Cityscapes dataset (See Table \ref{table:gta2city}). The Cityscapes dataset contains 5,000 densely labeled RGB images of size $2040\times 1016$ from 27 different cities.  Here the task is pixel level semantic segmentation. Following the experiment in \cite{hoffman2016fcns}, we use the GTA5 images as the labeled source dataset and the Cityscapes images as the unlabeled target domain.

We point out that the convolutional networks in our model are interchangeable. We include results using a dilated ResNet encoder for fair comparison with previous work, but we found from our experiments that the best performance was achieved by using our new Dilated Densely-Connected Networks (i.e. Dilated DenseNets) for the encoders which are derived by replacing strided convolutions with dilated convolutions \cite{yu2017dilated} in the DenseNet architecture \cite{huang2016densely}. DenseNets have previously been used for image segmentation~\cite{jegou2017one} but their encoder/decoder structure is more cumbersome than what we proposed. We use a series of transposed convolutional layers for the decoders. For the discriminators we follow the work by  by Zhu et al. \cite{zhu2017unpaired} and use a few convolutional layers.

Due to computational and memory constraints, we down sample all images by a factor of two prior to feeding them into the networks. Output segmentations are bilinearly up sampled to the original resolution. We train our network on 256x256 patches of the down sampled images, but test on the full images convolutionally. Furthermore, we did not include the cycle consistency constraint as that would require an additional pass through the encoder and decoder for both source and target images. Although cycle consistency regularizes the mappings more, we found that the identity and translation losses alone are enough in this case due to our shared latent space.

\begin{figure*}
\centering
\includegraphics[width=.99\linewidth]{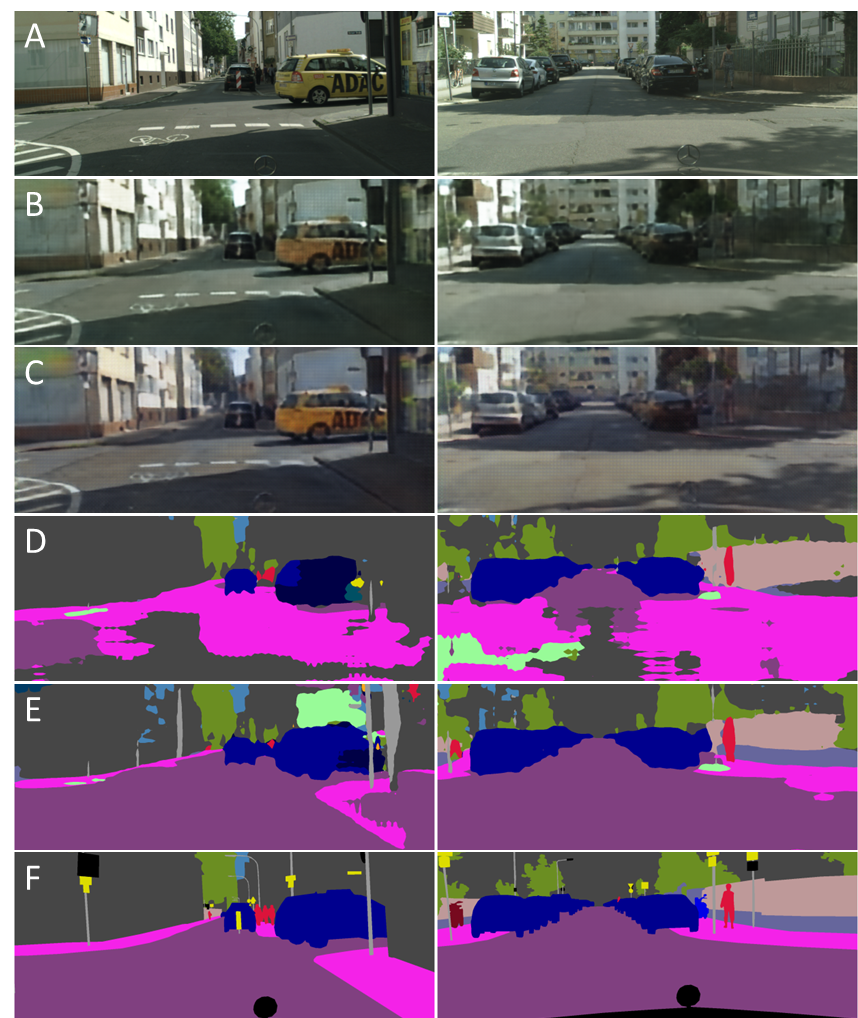}
\caption{A) Input image from real Cityscapes dataset. B) Identity mapped image. C) Translated image. D) Segmentation without domain adaptation. E) Our Segmentation. F) Ground truth. Although our image translations might not be as visually pleasing as those in \cite{zhu2017unpaired} (our architecture is not optimized for translation), they succeed in their goal of domain adaptation.}
\label{fig:gta2city}
\end{figure*}

Our encoder architecture (dilated ResNet/DenseNet) is optimized for segmentation and thus it is not surprising that our translations (see Figure.~\ref{fig:gta2city}) are not quite as good as those reported in \cite{zhu2017unpaired}. Qualitatively, it can be seen from Figure \ref{fig:gta2city} that our segmentations are much cleaner compared to no adaptation. Quantitatively (see Table~\ref{table:gta2city}), our method outperforms the previous method \cite{hoffman2016fcns} on all categories except 3, and is $5\%$ better overall. Further more, we show that using Dilated DenseNets in our framework, increases the SOA by $8.6\%$.


\begin{table*}
\setlength{\tabcolsep}{0.25em}
\small
\centering
\begin{tabular}{ c c c c c c c c c c c c c c c c c c c c c c} 
\hline
Method &
{\rotatebox[origin=c]{90}{road}} &
{\rotatebox[origin=c]{90}{~~sidewalk~~}} &
{\rotatebox[origin=c]{90}{building}} & 
{\rotatebox[origin=c]{90}{wall}} & 
{\rotatebox[origin=c]{90}{fence}} &
{\rotatebox[origin=c]{90}{pole}} &
{\rotatebox[origin=c]{90}{t light}} &
{\rotatebox[origin=c]{90}{t sign}} &
{\rotatebox[origin=c]{90}{veg}} &
{\rotatebox[origin=c]{90}{terrain}} &
{\rotatebox[origin=c]{90}{sky}} &
{\rotatebox[origin=c]{90}{person}} &
{\rotatebox[origin=c]{90}{rider}} &
{\rotatebox[origin=c]{90}{car}} &
{\rotatebox[origin=c]{90}{truck}} &
{\rotatebox[origin=c]{90}{bus}} &
{\rotatebox[origin=c]{90}{train}} &
{\rotatebox[origin=c]{90}{mbike}} &
{\rotatebox[origin=c]{90}{bike}} &
mIoU &\\ 
\hline
Source only &31.9&18.9&47.7&7.4&3.1&16.0&10.4&1.0&\textcolor{blue}{76.5}&13.0&58.9&36.0&1.0&67.1&9.5&3.7&0.0&0.0&0.0&21.1 \\ 
FCNs in the Wild~\cite{hoffman2016fcns} &67.4&29.2&64.9&15.6&8.4&12.4&9.8&2.7&74.1&12.8&\textcolor{blue}{66.8}&38.1&2.3&63.0&9.4&5.1&0.0&3.5&0.0&27.1  \\ 

I2I Adapt (Ours) &\textcolor{blue}{85.3}&\bf{38.0}&\textcolor{blue}{71.3}&\textcolor{blue}{18.6}&\textcolor{blue}{16.0}&\textcolor{blue}{18.7}&\textcolor{blue}{12.0}&\textcolor{blue}{4.5}&72.0&\textcolor{blue}{43.4}&63.7&\textcolor{blue}{43.1}&\textcolor{blue}{3.3}&\textcolor{blue}{76.7}&\textcolor{blue}{14.4}&\textcolor{blue}{12.8}&\textcolor{blue}{0.3}&\textcolor{blue}{9.8}&\textcolor{blue}{0.6}&\textcolor{blue}{31.8}  \\ 
\hline
Source only - DenseNet &67.3&23.1&69.4&13.9&14.4&21.6&\bf{19.2}&\bf{12.4}&78.7&24.5&74.8&49.3&3.7&54.1&8.7&5.3&2.6&6.2&\bf{1.9}&29.0 \\ 
I2I Adapt - DenseNet (Ours) &\bf{85.8}&37.5 &\bf{80.2}&\bf{23.3}&\bf{16.1}&\bf{23.0}&14.5&9.8&\bf{79.2}&\bf{36.5}&\bf{76.4}&\bf{53.4}&\bf{7.4}&\bf{82.8}&\bf{19.1}&\bf{15.7}&\bf{2.8}&\bf{13.4}&1.7&\bf{35.7} \\ 
\hline
\end{tabular}
\caption{Performance (Intersection over Union) of various methods on driving datasets domain adaptation. Above the line uses the standard dilated ResNet as the encoder. Our method performs the best overall and on all sub categories except two. Switching to a DenseNet encoder beats the previous method even without domain adaptation. DenseNet plus our method significantly out performs the previous method. Blue is best with ResNet, Bold is best overall.}
\label{table:gta2city}
\end{table*}

\section{Implementation Details}
What follows are further details about our network architectures, hyperparameters, and training procedures. We also plan to release our code with the conference version of the paper.

\subsection{MNIST, USPS, and SVHN digits datasets}
All images from MNIST and USPS were bilinearly upsampled to 32x32. Images from SVHN were converted to gray scale. All images were normalized to $[-1,1]$.

Our modified LeNet encoder consists of 4 stride 2 convolutional layers with 4x4 filters and 64, 64, 128, 128 features respectively. Each convolution is followed by batch normalization and a ReLU nonlinearity. Batch normalization was helpful with the image to image translation training. All weights are shared between the source and target encoders. Our DenseNet encoder follows~\cite{huang2016densely} with the final fully connected layer removed.

Our decoder consists of 4 stride 2 transposed convolutional layers with 4x4 filters and 512, 256, 128, 1 features respectively. Each convolution is followed by batch normalization and a ReLU nonlinearity except the last layer which only has a Tanh nonlinearity. The weights of the first two layers are shared between the source and target decoders.

The feature discriminator consists of 3 linear layers with 500, 500, 1 features, each followed by a leaky ReLU nonlinearity with slope $0.2$. The feature discriminator is trained with the Least Squares GAN loss. The Loss is only backpropagted to the generator for target images (we want the encoder to learn to map the target images to the same distribution as the source images, not vice versa).

The image discriminators consist of 4 stride 2 convolutional layers with 4x4 filters and 64, 128, 256, 1 features respectively. Each convolution is followed by instance normalization and a leaky ReLU nonlinearity with slope $0.2$. The image discriminators are trained with the Improved Waserstien loss with a gradient penalty of $10.0$.

For our hyperparameters we used:
$\lambda_c=1.0$,
$\lambda_{z}=0.2$,
$\lambda_{tr}=0.02$,
$\lambda_{id}=0.1$,
$\lambda_{cyc}=0.05$,
$\lambda_{trc}=0.0$.
The networks are trained using the ADAM optimizer with learning rate $0.0002$ and betas $0.5$ and $0.999$.

The translation classification loss ($Q_{trc}$) does not help with these simple digits datasets because the decoders can easily learn a permutation of the digits (for example a 2 may be translated to an 8 and then translated back to a 2).

\subsection{Office dataset}
Images are down sampled to 256x256 and then a random crop of size 224x244 is extracted.

For our encoder we use a ResNet34 pretrained on ImageNet. We do not use any dilation and thus have an output stride of 32. The final classification layer is applied after global average pooling.

Our decoders consist of a 5 4x4 stride 2 transposed convolutional layers with feature dimension 512, 256, 128, 64, 3. Each convolution is followed by batch normalization and a ReLU nonlinearity except the last layer which only has a Tanh nonlinearity. The weights of the first two layers are shared between the source and target decoders. We use the same image discriminator as in GTA5 to Cityscapes.
Here we found that using the Least Squares GAN loss produced better results.

The feature discriminator consists of 3 1x1 convolution layers with 500, 500, 1 features, each followed by a leaky ReLU nonlinearity with slope $0.2$.

Our hyperparameters were:
$\lambda_c=1.0$,
$\lambda_{z}=0.1$,
$\lambda_{tr}=0.005$,
$\lambda_{id}=0.2$,
$\lambda_{cyc}=0.0$,
$\lambda_{trc}=0.1$.
The networks are trained using the ADAM optimizer with learning rate $0.0002$ and betas $0.5$ and $0.999$. However, the pretrained encoder is trained with a learning rate of $5\times 10^{-5}$, to keep the weights closer to their good initialization.

\subsection{GTA5 to Cityscapes}
During training we use 512x512 crops which are down sampled by a factor of two prior to feeding them into the nets. Segmentation results are bilinearly up sampled to the full resolution. At test time we compute segmentations of the entire image convolutionally.

For our encoders we use a dilated ResNet34 and a dilated DenseNet121. We dilate the final two layers (blocks) so that the networks have an output stride of 8. We initialize the weights using a pretrained ImageNet classifier. Following \cite{yu2017dilated} we also add a dilation 2 followed by dilation 1 convolution layer to the end of the network to remove checkerboarding artifacts (whose weights are randomly initialized).

Our decoders consist of a 3x3 stride 1 convolutional layer followed by 3 4x4 stride 2 transposed convolutional layers with feature dimension 512, 256, 128, 3. Each convolution is followed by batch normalization and a ReLU nonlinearity except the last layer which only has a Tanh nonlinearity. The weights of the first two layers are shared between the source and target decoders.

The image discriminators consist of 4 stride 2 convolutional layers with 4x4 filters and 64, 128, 256, 1 features respectively. Each convolution is followed by instance normalization and a leaky ReLU nonlinearity with slope $0.2$. The image discriminators are trained with the Improved Waserstien loss with a gradient penalty of $10.0$. We did not use the feature discriminator for this experiment.

Although cycle consistency provides further regularization, it is computationally expensive for large images. We found that the identity and translation losses were enough to constrain the feature space.

Our hyperparameters were:
$\lambda_c=1.0$,
$\lambda_{z}=0.0$,
$\lambda_{tr}=0.04$,
$\lambda_{id}=0.2$,
$\lambda_{cyc}=0.0$,
$\lambda_{trc}=0.1$.
The networks are trained using the ADAM optimizer with learning rate $0.0002$ and betas $0.5$ and $0.999$.

The translated classification loss ($Q_{trc}$) is only backpropagated through the second encoding step ($f_y$). This prevents $f_x$ and $g_y$ from cheating and hiding information in the translated images to help $f_y$.

\section{Conclusion}
We have proposed a general framework for unsupervised domain adaptation which encompasses many recent works as special cases. Our proposed method simultaneously achieves image to image translation, source discrimination, and domain adaptation.

Our implementation outperforms state of the art on adaptation for digit classification and  semantic segmentation of driving scenes. When combined with the DenseNet architecture our method significantly outperforms the current state of the art.

\section*{Acknowledgments}\vspace{-0.2cm}
We gratefully acknowledge the support of NVIDIA Corporation with the donation of a Titan X Pascal GPU used for this research.

\bibliographystyle{ieee}
\bibliography{DAI2I}

\end{document}